\documentclass{Interspeech2024}

\usepackage[table,dvipsnames]{xcolor}
\usepackage{todonotes}
\usepackage{amsmath,amssymb,graphicx}
\usepackage{tabularx}
\usepackage{multirow}
\usepackage{ctable,booktabs}
\usepackage[inline]{enumitem}
\usepackage{makecell}
\usepackage{subcaption}

\usepackage[
backend=biber,
style=ieee,
citestyle=numeric-comp,
maxbibnames=3,
maxcitenames=3,
doi=false,isbn=false,url=false,eprint=false
]{biblatex}
\defbibheading{bibliography}[\refname]{}

\DeclareSourcemap{
    \maps[datatype=bibtex, overwrite=true]{
        \map{
            \step[fieldsource=booktitle,
            match=\regexp{.*Interspeech.*},
            replace={Proc. Interspeech}]
            \step[fieldsource=journal,
            match=\regexp{.*INTERSPEECH.*},
            replace={Proc. Interspeech}]
            \step[fieldsource=booktitle,
            match=\regexp{.*ICASSP.*},
            replace={Proc. ICASSP}]
            \step[fieldsource=booktitle,
            match=\regexp{.*icassp_inpress.*},
            replace={Proc. ICASSP (in press)}]
            \step[fieldsource=booktitle,
            match=\regexp{.*Acoustics,.*Speech.*and.*Signal.*Processing.*},
            replace={Proc. ICASSP}]
            \step[fieldsource=booktitle,
            match=\regexp{.*International.*Conference.*on.*Learning.*Representations.*},
            replace={Proc. ICLR}]
            \step[fieldsource=booktitle,
            match=\regexp{.*International.*Conference.*on.*Computational.*Linguistics.*},
            replace={Proc. COLING}]
            \step[fieldsource=booktitle,
            match=\regexp{.*SIGdial.*Meeting.*on.*Discourse.*and.*Dialogue.*},
            replace={Proc. SIGDIAL}]
            \step[fieldsource=booktitle,
            match=\regexp{.*International.*Conference.*on.*Machine.*Learning.*},
            replace={Proc. ICML}]
            \step[fieldsource=booktitle,
            match=\regexp{.*North.*American.*Chapter.*of.*the.*Association.*for.*Computational.*Linguistics:.*Human.*Language.*Technologies.*},
            replace={Proc. NAACL}]
            \step[fieldsource=booktitle,
            match=\regexp{.*Empirical.*Methods.*in.*Natural.*Language.*Processing.*},
            replace={Proc. EMNLP}]
            \step[fieldsource=booktitle,
            match=\regexp{.*Association.*for.*Computational.*Linguistics.*},
            replace={Proc. ACL}]
            \step[fieldsource=booktitle,
            match=\regexp{.*Automatic.*Speech.*Recognition.*and.*Understanding.*},
            replace={Proc. ASRU}]
            \step[fieldsource=booktitle,
            match=\regexp{.*Spoken.*Language.*Technology.*},
            replace={Proc. SLT}]
            \step[fieldsource=booktitle,
            match=\regexp{.*Speech.*Synthesis.*Workshop.*},
            replace={Proc. SSW}]
            \step[fieldsource=booktitle,
            match=\regexp{.*workshop.*on.*speech.*synthesis.*},
            replace={Proc. SSW}]
            \step[fieldsource=booktitle,
            match=\regexp{.*Advances.*in.*neural.*information.*processing.*},
            replace={Proc. NeurIPS}]
            \step[fieldsource=booktitle,
            match=\regexp{.*Advances.*in.*Neural.*Information.*Processing.*},
            replace={Proc. NeurIPS}]
            \step[fieldsource=booktitle,
            match=\regexp{.*Workshop.*on.* Applications.* of.* Signal.*Processing.*to.*Audio.*and.*Acoustics.*},
            replace={Proc. WASPAA}]
            \step[fieldsource=publisher,
            match=\regexp{.+},
            replace={{}}]
            \step[fieldsource=month,
            match=\regexp{.+},
            replace={{}}]
            \step[fieldsource=location,
            match=\regexp{.+},
            replace={{}}]
            \step[fieldsource=address,
            match=\regexp{.+},
            replace={{}}]
            \step[fieldsource=organization,
            match=\regexp{.+},
        replace={{}}]
        }
    }
}

\newcommand{\sysname}{\textsc{Multi-Convformer}}
\newcommand{\blockname}{\textsc{MultiConv}}
\newcolumntype{P}[1]{>{\centering\arraybackslash}p{#1}}

\interspeechcameraready

\title{\sysname: Extending Conformer with \\ Multiple Convolution Kernels}

\name[affiliation={1}]{Darshan}{Prabhu}
\name[affiliation={2}]{Yifan}{Peng}
\name[affiliation={1}]{Preethi}{Jyothi}
\name[affiliation={2}]{Shinji}{Watanabe}

\address{
  $^1$Department of Computer Science, Indian Institute of Technology Bombay, Mumbai, India \\
  $^2$Language Technologies Institute, Carnegie Mellon University, Pittsburgh, PA, USA
}
\email{\{darshanp, pjyothi\}@cse.iitb.ac.in, yifanpen@andrew.cmu.edu, shinjiw@ieee.org}


\begin{document}

\maketitle

\begin{abstract}
Convolutions have become essential in state-of-the-art end-to-end Automatic Speech Recognition~(ASR) systems due to their efficient modelling of local context. Notably, its use in Conformers has led to superior performance compared to vanilla Transformer-based ASR systems. While components other than the convolution module in the Conformer have been reexamined, altering the convolution module itself has been far less explored. Towards this, we introduce \sysname~ that uses multiple convolution kernels within the convolution module of the Conformer in conjunction with gating. This helps in improved modeling of local dependencies at varying granularities. Our model rivals existing Conformer variants such as CgMLP and E-Branchformer in performance, while being more parameter efficient. We empirically compare our approach with Conformer and its variants across four different datasets and three different modelling paradigms and show up to 8\% relative word error rate~(WER) improvements.

\end{abstract}

\section{Introduction}
In recent years, end-to-end automatic speech recognition (ASR) systems have emerged as the preferred model of choice to obtain state-of-the-art ASR performance. An integral component of such systems is the speech encoder~\cite{e2e_asr_survey1,e2e_asr_survey2} that uses Attention via Transformers~\cite{attention} to map the input speech into high-level acoustic representations. 
Transformer-based ASR systems~\cite{speech_transformer2,speech_transformer3} often struggle to model local relationships effectively. 
To address this limitation, Gulati et~al.~\cite{conformer} proposed the Conformer~\cite{conformer_impl} architecture, that combines multi-headed attention with convolutions~\cite{convolution,convolution2,convolution3}. 
With the advent of Conformer models, the idea of using convolutions alongside attention to independently model both local and global relationships has been widely explored~\cite{branchformer, e_branchformer, zipformer, leformer}.


\looseness=-1
\begin{figure}[t]
  \centering
  \includegraphics[keepaspectratio, width=0.46\textwidth]{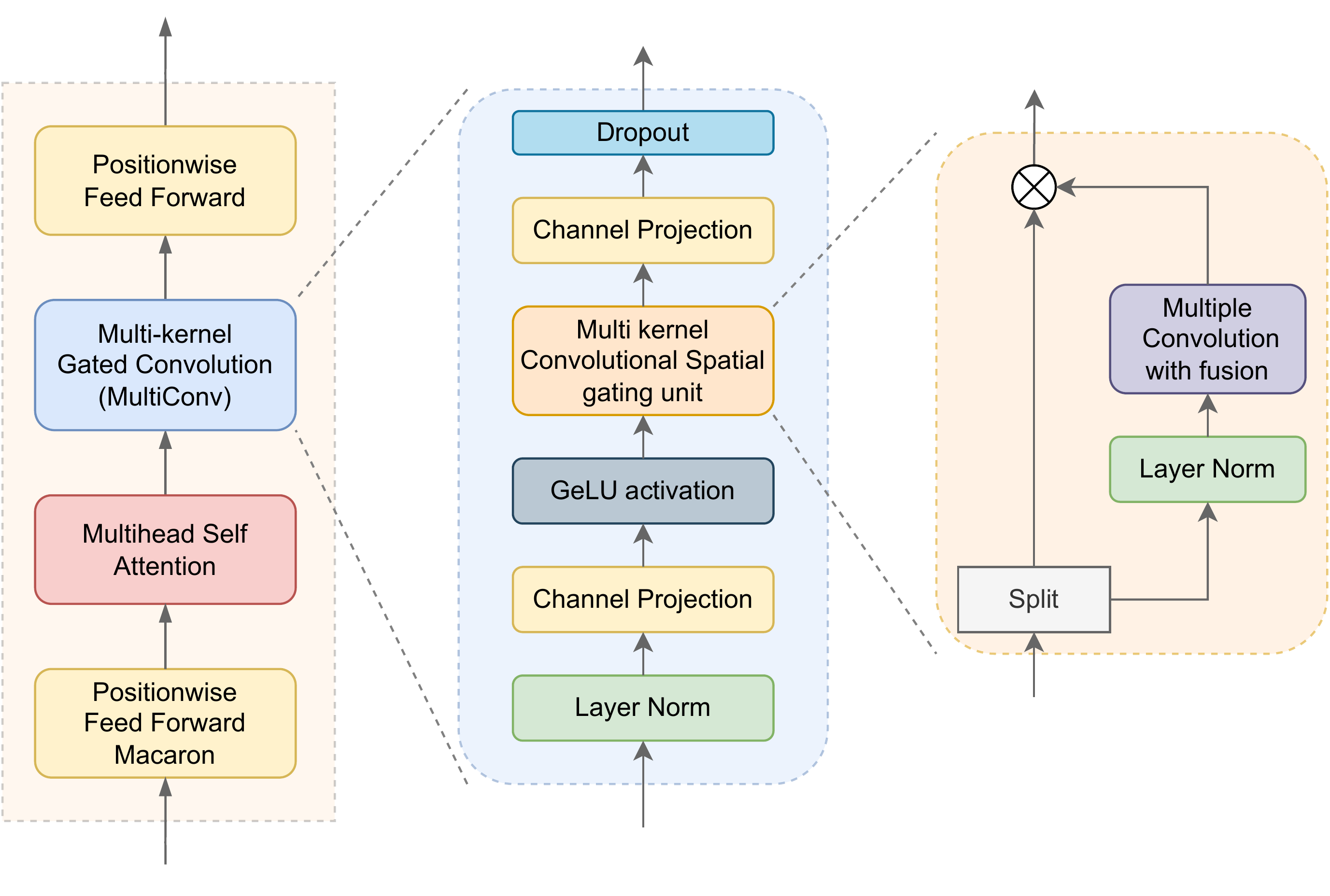}
  \caption{Overview of our \blockname~encoder layer. It comprises a stacked architecture similar to Conformer, except the convolution block is replaced with a gated multi-kernel convolution block. We omit residual connections and layer norm for readability. \looseness=-1 }  
  \vspace{-0.2cm}
  \label{fig:main_arch} 
\end{figure}

Despite these notable advancements, prior work~\cite{branchformer} has shown that the use of fixed-kernel convolutions within these models creates a bottleneck, forcing the model to re-purpose some of its attention heads to function as local information extractors. This negatively impacts the performance of attention, whose primary purpose is to model global information. To address this limitation, in our work, we propose a multiple convolution-based enhancement to the Conformer architecture (that we call \sysname). 
Additionally, we incorporate gating, a technique that has proven to be effective in encoder architectures~\cite{cgmlp,branchformer,e_branchformer}. By combining these approaches, we achieve significant improvements~(up to 8\% relative WER improvement) over the original Conformer architecture and perform at par or better than its variants such as CgMLP~\cite{cgmlp} and E-Branchformer~\cite{e_branchformer}. 
The use of multiple convolutions in order to generate better local context has been widely adopted in image-related tasks~\cite{dynamic_convolution,dynamic_convolution2,dynamic_convolution3,dynamic_convolution4,dynamic_convolution5}. Such enhancements have also been used in speech emotion recognition~\cite{ser} and robust ASR~\cite{multi_octave,multi_stream}; the latter works use multiple convolutions within fully convolutional or TDNN-style architectures (unlike our work).


\looseness=-1
\begin{figure*}[t]
  \centering
  \includegraphics[keepaspectratio, width=\textwidth]{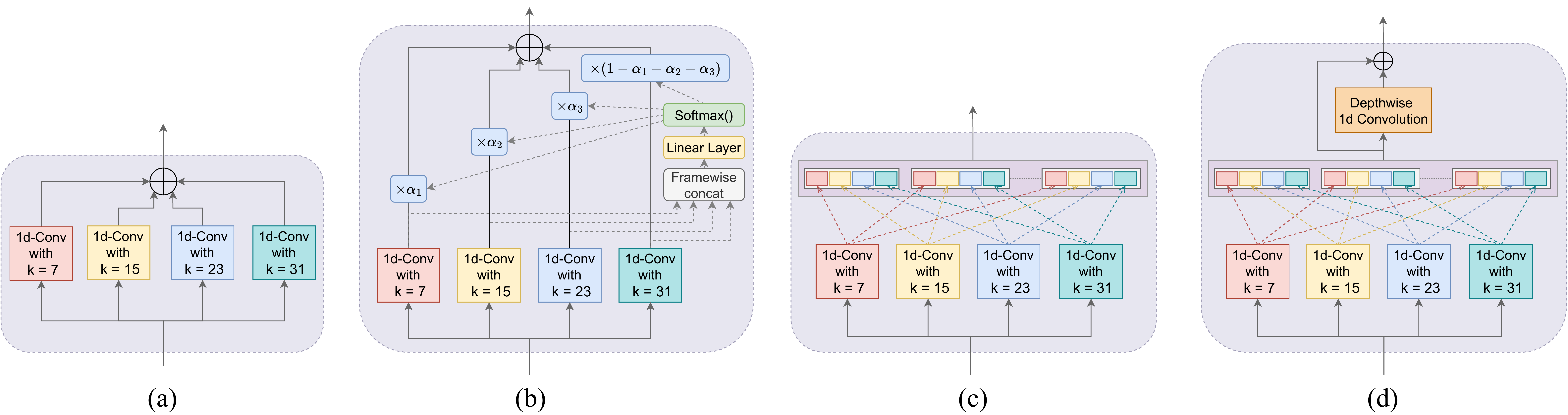}
  
  \caption{Overview of the four variants of the \textsc{Fusion}() operation in our proposed \textbf{Multi-kernel Convolutional Spatial Gating Unit}~(\texttt{M-CSGU}) block that demonstrates how convolution kernels with $K=\{7,15,23,31\}$ are used in our \blockname~block. Briefly, the four architectures are as follows: In \textbf{(a)} element-wise addition is employed on the outputs of the convolutions to generate the final output. \textbf{(b)} builds upon (a) by learning weights that determine the importance of each convolution's output. \textbf{(c)} uses compression kernels to reduce the input by $1/4^{\text{th}}$ and then concatenates all of them frame-wise to generate the output. \textbf{(d)} further enhances (c) by using an additional depthwise convolution at the end that takes neighboring frames into account when generating the final output.}   
  \label{fig:arch:fusion} 
  \vspace{-0.2cm}
\end{figure*}

\noindent In summary, our main contributions are as follows:

\begin{itemize}
    \item We propose \sysname, a variant of Conformer that uses multiple convolution kernels instead of a fixed kernel convolution to capture local context more effectively.%
    \footnote{Our code is available in the \href{https://github.com/espnet/espnet}{ESPnet toolkit}.}
    \item We show the effectiveness of our approach by comparing with Conformer and its variants on ASR and Spoken Language Understanding~(SLU). We experiment with multiple datasets~(Librispeech~\cite{librispeech}, Tedlium2~\cite{tedlium2}, AISHELL~\cite{aishell} and SLURP~\cite{slurp}) and various modelling paradigms and obtain up to 8\% relative WER improvement over Conformer. We also conduct several analyses and ablations to showcase the effectiveness and interpretability of our approach.
\end{itemize}

\section{Methodology}

In this work, we experiment with the three most popular ASR architectures:~the attention-based encoder-decoder model~(AED)~\cite{jointctcatt}, the encoder-only model~(pure CTC)~\cite{peng2024owsmctc} and the RNN-Transducer model~(RNN-T)~\cite{rnnt}. An essential component common to all these architectures is the encoder~(\textsc{Enc}) module that maps an input sequence of speech features$X=\{ \mathbf{x}_1, \mathbf{x}_2, \ldots, \mathbf{x}_{L} | \mathbf{x}_i \in \mathbb{R}^{c} \}$ to a (typically smaller) sequence of contextualized representations $H=\textsc{Enc}(X) = \{ \mathbf{h}_1, \mathbf{h}_2, \ldots \mathbf{h}_T | \mathbf{h}_i \in \mathbb{R}^{d}  \} $.
Thereafter, the manner in which $H$ is trained to predict the final $M$-length ground-truth token sequence $Y=\{y_1,y_2,\ldots, y_M | y_i \in \mathbb{N}^+ \}$ depends on the underlying architecture. AED uses an attention-based decoder and a CTC module to jointly learn a mapping from $H$ to $Y$. However, pure CTC and RNN-T are encoder-only models that employ CTC~\cite{ctc} and RNN-T~\cite{rnnt} losses, respectively, to generate $Y$.
Since our modifications are constrained to the encoder, in subsequent sections, we restrict our discussion only to the composition of the \textsc{Enc} module. 
\looseness=-1

\subsection{\sysname~Encoder}\label{subsec:arch:backbone}
Figure~\ref{fig:main_arch} illustrates the overall architecture of a single \sysname\ encoder layer and Figure~\ref{fig:arch:fusion} gives an overview of how the outputs from multiple convolutions are merged together. The \sysname~encoder layer consists of four blocks that are stacked together and interspersed with layer normalization~\cite{layernorm} and residual connections. The two position-wise feed-forward layers aid in refining the point-wise information, while the multi-head attention and convolution are responsible for incorporating contextual information. This stacked architecture has been widely used in prior work~\cite{conformer, branchformer, e_branchformer, squeezeformer}. We adopt this same stack, but replace the fixed single-kernel convolution block with a more expressive multi-kernel convolution module that will henceforth be referred to as \blockname. 
\looseness=-1

\looseness=-1

\begin{table*}[t!]
\caption{Comparing the performance~(CER or WER \%) of our proposed system against Transformer and Conformer on three datasets. The three sections are: (1) AED: Encoder-Decoder models with joint CTC-Attention loss, (2) Pure CTC: Encoder-only model where only CTC loss is employed and (3) RNN-T: RNN based Transducer model. \colorbox{green!20}{$\phantom{x}$} denotes the best CER / WER across all experiments. $\dagger$ and $\ddagger$ indicate statistically significant results compared to Conformer at $p<0.05$ and $p<0.001$ using MAPSSWE test~\cite{mapsswe}, respectively. \looseness=-1 }
  \vspace{-0.2cm}
\label{table:res_main}
\centering
\resizebox{\linewidth}{!}{
    \begin{tabular}{ l|c|cccc|c|cc|c|cc }
        \hline
        \hline
        
        \multicolumn{1}{c|}{\multirow{2}{*}{\textbf{\small{Method}}}} & \multicolumn{5}{c|}{\textbf{\footnotesize{Librispeech-100h~(WER)}}} & \multicolumn{3}{c|}{\textbf{\footnotesize{Tedlium2~(WER)}}} & \multicolumn{3}{c}{\textbf{\footnotesize{AISHELL~(CER)}}} \\
        
        \cline{2-12}

        & \textbf{\scriptsize{\# params}} & \textbf{\scriptsize{Dev Clean}} & \textbf{\scriptsize{Dev Other}} & \textbf{\scriptsize{Test Clean}} & \textbf{\scriptsize{Test Other}} & 
        
        \textbf{\scriptsize{\# params}} & \textbf{\scriptsize{Dev}} & \textbf{\scriptsize{Test}} & 
        
        \textbf{\scriptsize{\# params}} & \textbf{\scriptsize{Dev}} & \textbf{\scriptsize{Test}}  \\
        
        \hline
        \multicolumn{12}{c}{\footnotesize{Attention Encoder Decoder~(AED) models}} \\
        \hline
        
        \small{Transformer \cite{speech-transformer}} & \footnotesize{40.9M} &  \footnotesize{8.02} &  \footnotesize{20.14} &  \footnotesize{8.39} & \footnotesize{20.34} & \footnotesize{33.8M} & \footnotesize{10.13} & \footnotesize{8.83} & \footnotesize{47.7M} & \footnotesize{5.16} & \footnotesize{5.53} \\
    
        \small{Conformer \cite{conformer}} & \footnotesize{37.4M} &  \footnotesize{6.59} &  \footnotesize{17.24} &  \footnotesize{6.89} & \footnotesize{17.27} & \footnotesize{33.9M} & \footnotesize{9.30} & \footnotesize{7.65} & \footnotesize{54.2M} & \footnotesize{4.23} & \footnotesize{4.63} \\
        

        \arrayrulecolor{black!50}
        \hline
        \arrayrulecolor{black}

        \small{$\blockname_{\texttt{sum}}$} & 
        \footnotesize{37.0M} &  \footnotesize{6.00} &  \cellcolor{green!20} \footnotesize{\textbf{16.56}}~$\ddagger$ &  \footnotesize{6.33} &  \cellcolor{green!20} \footnotesize{\textbf{16.60}}~$\ddagger$ & 
        \footnotesize{33.5M} &  \cellcolor{green!20} \footnotesize{\textbf{7.99}}~$\ddagger$ & \footnotesize{7.34} & \footnotesize{54.2M} & \footnotesize{4.18} & \footnotesize{4.47} \\

        \small{$\blockname_{\texttt{weighted}}$} & \footnotesize{37.1M} &  \footnotesize{6.17} &  \footnotesize{17.03} &  \footnotesize{6.48} & \footnotesize{17.36} &
        \footnotesize{33.6M} & \footnotesize{8.15} & \footnotesize{7.41} & \footnotesize{54.4M} & \footnotesize{4.40} & \footnotesize{4.67} \\

        \small{$\blockname_{\texttt{concat}}$} & \footnotesize{37.0M} &  \footnotesize{6.23} &  \footnotesize{17.19} &  \footnotesize{6.41} & \footnotesize{17.15} & 
        \footnotesize{33.5M} & \footnotesize{8.32} & \footnotesize{7.42} & \footnotesize{54.2M} & \cellcolor{green!20} \footnotesize{\textbf{4.16}}~$\dagger$ & \cellcolor{green!20} \footnotesize{\textbf{4.46}}~$\dagger$ \\

        \small{$\blockname_{\texttt{depth}}$} & \footnotesize{37.2M} &   \cellcolor{green!20} \footnotesize{\textbf{5.87}}~$\ddagger$ &  \footnotesize{16.63} & \cellcolor{green!20} \footnotesize{\textbf{6.18}}~ $\ddagger$ & \footnotesize{17.00} & 
        \footnotesize{33.7M} & \footnotesize{8.01} &  \cellcolor{green!20} \footnotesize{\textbf{7.27}}~$\ddagger$ & \footnotesize{54.7M} & \footnotesize{4.18} &  \cellcolor{green!20} \footnotesize{\textbf{4.46}} \\
    
        \hline
        \multicolumn{12}{c}{\footnotesize{Encoder Only~(Pure CTC) models}} \\
        \hline

        \small{Transformer \cite{speech-transformer}} & \footnotesize{28.9M} &  \footnotesize{ 12.32 } &  \footnotesize{ 27.49 } &  \footnotesize{ 12.88 } & \footnotesize{ 28.26 } & \footnotesize{27.7M} & \footnotesize{ 11.97 } & \footnotesize{ 11.71 } & \footnotesize{28.7M} & \footnotesize{ 6.51 } & \footnotesize{ 7.02 } \\

        \small{Conformer \cite{conformer}} & \footnotesize{25.4M} &  \footnotesize{9.33} &  \footnotesize{22.60} &  \footnotesize{9.76} & \footnotesize{23.10} & 
        \footnotesize{24.2M} & \footnotesize{9.65} & \footnotesize{8.80} & \footnotesize{39.9M} & \footnotesize{5.98} & \footnotesize{6.59} \\


        \arrayrulecolor{black!50}
        \hline
        \arrayrulecolor{black}

        \small{$\blockname_{\texttt{depth}}$} & \footnotesize{25.1M} & \cellcolor{green!20} \footnotesize{\textbf{9.24}} &  \cellcolor{green!20} \footnotesize{\textbf{22.26}}~$\dagger$ & \cellcolor{green!20} \footnotesize{\textbf{9.47}}~$\dagger$ & \cellcolor{green!20} \footnotesize{\textbf{23.09}} &
        \footnotesize{24.0M} & \cellcolor{green!20} \footnotesize{\textbf{8.81}}~$\dagger$  & \cellcolor{green!20} \footnotesize{\textbf{8.44}}~$\ddagger$  & \footnotesize{40.2M} & \cellcolor{green!20} \footnotesize{\textbf{5.67}}~$\ddagger$ & \cellcolor{green!20} \footnotesize{\textbf{6.01}}~$\ddagger$ \\

        \hline
        \multicolumn{12}{c}{\footnotesize{RNN Transducer~(RNN-T) models}} \\
        \hline

        \small{Transformer \cite{speech-transformer}} & \footnotesize{32.5M} &  \footnotesize{ 9.18 } &  \footnotesize{ 22.40 } &  \footnotesize{ 9.43 } & \footnotesize{ 22.95 } & 
        \footnotesize{28.7M} & \footnotesize{ 10.26 } & \footnotesize{ 9.65 } & \footnotesize{31.8M} & \footnotesize{ 5.69 } & \footnotesize{ 6.21 } \\

        \small{Conformer \cite{conformer}} & \footnotesize{28.9M} &  \footnotesize{6.79} &  \footnotesize{18.28} &  \footnotesize{7.19} & \footnotesize{18.70} & 
        \footnotesize{25.2M} & \footnotesize{8.37} & \footnotesize{7.96} & \footnotesize{43.0M} & \footnotesize{4.93} & \footnotesize{5.22} \\


        \arrayrulecolor{black!50}
        \hline
        \arrayrulecolor{black}

        \small{$\blockname_{\texttt{depth}}$} & \footnotesize{28.7M} &  \cellcolor{green!20} \footnotesize{\textbf{6.60}} & \cellcolor{green!20} \footnotesize{\textbf{17.53}}~$\ddagger$ & \cellcolor{green!20} \footnotesize{\textbf{7.05}} & \cellcolor{green!20} \footnotesize{\textbf{18.23}}~$\ddagger$ & 
        \footnotesize{24.9M} & \cellcolor{green!20} \footnotesize{\textbf{8.20}} & \cellcolor{green!20} \footnotesize{\textbf{7.67}} & \footnotesize{43.4M} & \cellcolor{green!20} \footnotesize{\textbf{4.65}}~$\ddagger$ & \cellcolor{green!20} \footnotesize{\textbf{5.05}} \\
        
        \hline
        \hline
    \end{tabular}
}
\vspace{-0.4cm}
\end{table*}
As illustrated in Figure~\ref{fig:main_arch}, in a single encoder layer, the output from the multi-head attention block $A = \{\mathbf{a}_1,\mathbf{a}_2,\ldots,\mathbf{a}_T\ | \mathbf{a}_j \in \mathbb{R}^{d}\}$ is first normalized with a layer normalization. Subsequently, it undergoes channel projection to increase its dimensionality from $d$ to $d_{\text{inter}}$%
\footnote{Here $d_{\text{inter}} > d$. Typically, $d_{\text{inter}}=6d$. Using a higher intermediate dimension has been shown to be an effective strategy for position-wise feedforward layers~\cite{positionwise_ff}.}.
To introduce non-linearity to the representation, we apply \textsc{GELU}~\cite{gelu} activation. The resulting output, $\hat{A} = \{\mathbf{\hat{a}}_1,\mathbf{\hat{a}}_2,\ldots,\mathbf{\hat{a}}_T | \mathbf{\hat{a}}_j \in \mathbb{R}^{d_{\text{inter}}}\}$, is then passed through our Multi-kernel Convolutional Spatial Gating Unit~(\texttt{M-CSGU}). \texttt{M-CSGU} is a more powerful alternative~\cite{branchformer,e_branchformer} to the standard convolution block due to its usage of gates~\cite{gated_mlp} along with convolutions. Finally, we employ another channel projection layer that projects the output from $\mathbb{R}^{d_{\text{inter}}}$ back to $\mathbb{R}^{d}$, followed by dropout for regularization. 



\vspace{0.15cm}
\noindent \textbf{Multi-kernel Convolutional Spatial Gating Unit}~(\texttt{M-CSGU}): \texttt{M-CSGU} module takes $\hat{A}$ as its input. We first bifurcate each representation in $\hat{A}$ into two parts, each having dimension $d' = d_{\text{inter}}/2$. Only one part undergoes layer normalization and passes through multiple convolutions; the other part stays intact. These parts are multiplied element-wise, thus creating a gate. Next, we use $P$ depthwise convolutions with kernel sizes $K=\{k_1,k_2,\ldots,k_P\}$. Formally, these operations are as follows:
\looseness=-1

\begin{align}
    [ Z_l, Z_r ] &= [ \hat{A}[:,:d'], \texttt{LayerNorm}(\hat{A}[:,d':]) ] \nonumber \\
    [V_1, V_2, \ldots, V_P] &= [ \texttt{Conv}_{k_1}(Z_r), \ldots ,\texttt{Conv}_{k_P}(Z_r) ] \nonumber \\
    \tilde{Z}_r &= \textsc{Fusion}([V_1, V_2, \ldots, V_P]) \nonumber \\
    \hat{C} &= Z_l \odot \tilde{Z}_r 
    \label{eq:mcsgu}
\end{align}\looseness=-1
where $\hat{A}, \hat{C} \in \mathbb{R}^{T \times d_{\text{inter}}}$, $ Z_l, Z_r, V_j, \tilde{Z}_r \in \mathbb{R}^{T \times d'}$, $\odot$ represents element-wise products, $\texttt{Conv}_{k_i}()$ refers to a depthwise convolution with kernel size of $k_i$ and $\hat{C}$ is the final output from this block which is further passed to a position-wise feed-forward layer. Since $\textsc{Fusion}$ must preserve the dimensionality of the input, the number of input channels to each $\texttt{Conv}$ becomes $d'$, however the size of the output channels depends on the nature of the $\textsc{Fusion}()$ operation. We explore four different fusion mechanisms, shown in Figure~\ref{fig:arch:fusion}, that we can further group into two categories: Sum-based and Concat-based \textsc{Fusion}.

\vspace{0.15cm}
\noindent \textbf{Sum-based \textsc{Fusion}}: In this mechanism, both the input and output channels are of the same size. The outputs obtained from each convolution are combined using an element-wise addition operation as shown in Figure~\ref{fig:arch:fusion}(a). That is, $\tilde{Z}_r$ in Equation~\ref{eq:mcsgu} is computed as: $\tilde{Z}_r = V_1 + V_2 \ldots + V_P$. In our experiments, we refer to this fusion approach as $\blockname_{\texttt{sum}}$. We further enhance this fusion by learning weights that decide the importance of each convolution for every frame of the input, as shown in Figure~\ref{fig:arch:fusion}(b) and defined formally below:
\looseness=-1
\begin{align}
\alpha^s &= \{ \alpha^s_1, \ldots \alpha^s_P \} = \texttt{Softmax}( \textsc{FFN}_{d' \rightarrow P }(Z^s_r)) \nonumber \\
    \tilde{Z}^s_r &= \alpha^s_1 \cdot V^s_1 + \alpha^s_2 \cdot V^s_2 + \ldots + \alpha^s_P \cdot V^s_P \nonumber \\
    \tilde{Z}_r &= [\tilde{Z}^1_r, \tilde{Z}^2_r, \ldots \tilde{Z}^T_r] \nonumber
\end{align}
where $\textsc{FFN}_{a \rightarrow b}$ is a projection layer that projects $a$-dimensional inputs to $b$-dimensional outputs and $\alpha_j \in [0,1]$ is the importance given by the $s^{th}$ frame $Z^s_r$ to the output $V^s_j$ of the $j^{th}$ kernel. We refer to this fusion mechanism as $\blockname_{\texttt{weighted}}$ in our experiments. 
\looseness=-1

\vspace{0.15cm}
\noindent \textbf{Concat-based \textsc{Fusion}}: In contrast to sum, the concat-based fusion allocates a portion of the output to each convolution, causing the number of input and output channels for these convolutions to be different. This reconstruction allocates $1/P$ of the total number of features $d'$ to each convolution, resulting in the number of output channels to be $d'/P$ as shown in Figure~\ref{fig:arch:fusion}(c). This operation is shown below:
\begin{align}
    \tilde{Z}^s_r &= \texttt{Concat}(V^s_1 , V^s_2 \ldots , V^s_P) \nonumber \\ 
    \tilde{Z}_r &= [\tilde{Z}^1_r, \tilde{Z}^2_r, \ldots \tilde{Z}^T_r] \nonumber \label{eq:arch:concat} 
\end{align}
where $V^s_j, \tilde{Z}^s_r \in \mathbb{R}^{d'/P}$. In our experiments, we call this $\blockname_{\texttt{concat}}$.
We further enhance this fusion by introducing a depthwise convolution after the frame reconstruction to take neighboring frames into account while combining the convolution outputs, as shown in Figure~\ref{fig:arch:fusion}(d). We refer to this architecture as $\blockname_{\texttt{depth}}$ in our experiments.
\section{Experimental Setup}
We conduct experiments on five datasets namely Librispeech-100h~(LS-100)~\cite{librispeech}, Librispeech-960h~(LS-960)~\cite{librispeech}, Tedlium-2~\cite{tedlium2}, AISHELL-1~\cite{aishell} and SLURP~\cite{slurp}. We use ESPnet toolkit~\cite{espnet} to run all our experiments on a combination of NVIDIA RTX A6000 and NVIDIA Tesla V100 GPUs.%
\footnote{We ensure same GPU and environment is used while running all the experiments on a particular dataset.}
All our models take $80$-dimensional log-Mel features as input that are extracted with a 25ms window size and 10ms stride. We also use 3-way speed perturbation with ratios $\{ 0.9,1.0,1.1 \}$ and SpecAugment~\cite{specaug}. 
In all our experiments, we use the experimental settings recommended in ESPnet recipes. For all datasets except LS-960 and SLURP, the encoder-decoder architecture consists of $12$ encoder and $6$ decoder layers with an attention dimension of $d=256$ and $4$ attention heads. However, for LS-960 and SLURP, we use an $18$ layer encoder with $8$ attention heads and an attention dimension of $d=512$. 
\looseness=-1
\section{Experimental Results and Analysis}


Table~\ref{table:res_main} compares all four variants of our proposed \sysname~(elaborated in Section~\ref{subsec:arch:backbone}) to Transformer~\cite{speech-transformer} and Conformer~\cite{conformer} models.
Our method significantly outperforms both these approaches across all three datasets and multiple modelling paradigms; results that are statistically significant are shown with $\ddagger$. Additionally, we find that among the four \textsc{Fusion} methods~(Figure~\ref{fig:arch:fusion}), $\blockname_{\texttt{sum}}$ and $\blockname_{\texttt{depth}}$ perform better. Henceforth, we will only show results with using the $\blockname_{\texttt{depth}}$ fusion strategy.
\looseness=-1
\subsection{ASR and SLU Experiments}
\label{subsec:analysis:other_archs}

\begin{table}[t]
\caption{Comparison of our system~(WER \%) with CgConv and E-Branchformer~(reproduced using ESPnet recipes~\cite{espnet}) on the test splits of the three datasets.}
  \vspace{-0.05cm}

\label{table:res_branchformer}
\centering
\resizebox{\linewidth}{!}{
\begin{tabular}{ | l | cc | c | c | }
    \hline
    \multicolumn{1}{|c|}{\multirow{2}{*}{\textbf{\small{Method}}}} & \multicolumn{2}{c|}{\textbf{\footnotesize{Librispeech-100h}}} & \multicolumn{1}{c|}{\textbf{\footnotesize{Tedlium2}}} & \multicolumn{1}{c|}{\textbf{\footnotesize{AISHELL}}} \\
    
    \cline{2-5}
    
    & \textbf{\footnotesize{Test Clean}} & \textbf{\footnotesize{Test Other}} & \textbf{\footnotesize{Test}} &
    \textbf{\footnotesize{Test}} \\

    \hline
    \multicolumn{5}{|c|}{\footnotesize{Attention Encoder Decoder~(AED) models}} \\
    \hline

   \small{CgConv} & \small{6.36} & \small{17.18} & \small{7.44} & \small{4.55} \\

   \small{E-Branchformer}~\cite{e_branchformer} & \small{6.50} & \small{17.06} & \small{7.34} & \small{4.47} \\

   \arrayrulecolor{black!50}
    \hline
    \arrayrulecolor{black}

   \footnotesize{$\blockname_{\texttt{depth}}$} &  \small{\textbf{6.18}} & \small{\textbf{17.00}} & \small{\textbf{7.27}} & \small{\textbf{4.46}} \\

   \hline
    \multicolumn{5}{|c|}{\footnotesize{Encoder Only~(Pure CTC) models}} \\
    \hline

   \small{CgConv} & \small{9.66} & \small{23.79} & \small{8.78} & \small{6.33} \\

   \small{E-Branchformer}~\cite{e_branchformer} & \small{10.05} & \small{23.87} & \small{8.78} & \small{6.04} \\

   \arrayrulecolor{black!50}
    \hline
    \arrayrulecolor{black}

   \footnotesize{$\blockname_{\texttt{depth}}$} &  \small{\textbf{9.47}} & \small{\textbf{23.09}} & \small{\textbf{8.44}} & \small{\textbf{6.01}} \\

   \hline
    \multicolumn{5}{|c|}{\footnotesize{RNN Transducer~(RNN-T) models}} \\
    \hline

   \small{CgConv} & \small{6.95} & \small{18.18} & \small{7.72} & \small{5.24} \\

   \small{E-Branchformer}~\cite{e_branchformer} & \small{\textbf{6.87}} & \small{\textbf{18.09}} & \small{7.71} & \small{5.17} \\

   \arrayrulecolor{black!50}
    \hline
    \arrayrulecolor{black}

   \footnotesize{$\blockname_{\texttt{depth}}$} &  \small{7.05} & \small{18.23} & \small{\textbf{7.67}} & \small{\textbf{5.05}} \\
    
    \hline
\end{tabular}
}
\vspace{-0.35cm}
\end{table}

\begin{figure}[t]
  \centering
  \includegraphics[keepaspectratio, width=0.43\textwidth]{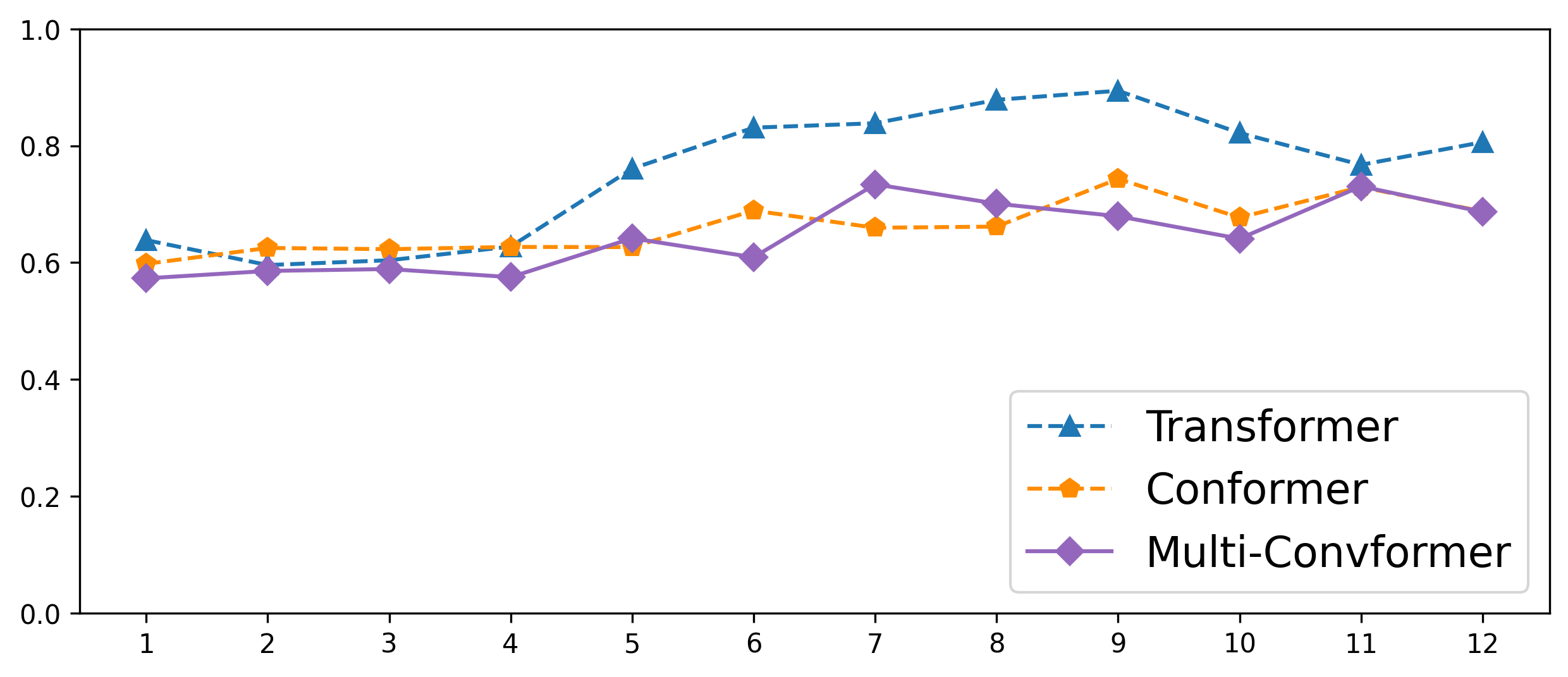}
  \caption{Line plot showing the degree of diagonality seen in each encoder layer's self-attention block. A higher value indicates that the attention weight matrix is more concentrated along its diagonal~\cite{attnetion_usefullness}. }   

  \label{fig:diagonality} 
\vspace{-0.1cm}
\end{figure}

\noindent \textbf{Comparison with Conformer variants.} Table~\ref{table:res_branchformer} shows the WER comparison between our system and two Conformer variants: CgConv~(Conformer with convolution replaced by Convolutional Spatial Gating Unit~\cite{cgmlp})%
\footnote{This is a stronger variant of the original CgMLP architecture~\cite{cgmlp}, which is a pure MLP-based system.}
and E-Branchformer~(Conformer with disentangled attention and convolution branches)~\cite{e_branchformer,comparative_study_branchformer}. 
We note here that $\blockname_{\texttt{sum}}$ can be considered as an improved version of CgConv, as it reduces to CgConv when a single fixed convolution kernel is employed instead of multiple kernels. Further, we find that the use of a gate in conjunction with convolution allows the model to selectively utilize convolution, thereby introducing a natural branching capability. As a result, our proposed architecture can be viewed as an enhancement to CgConv and a parameter-efficient variant of Branchformer that achieves comparable or better performance on speech-related tasks.

In Figure~\ref{fig:diagonality}, we compare the diagonal properties of self-attention blocks among Transformer, Conformer, and \sysname\ via the diagonality metric~\cite{attnetion_usefullness,branchformer}. This metric is an indication of the degree to which attention heads focus on capturing local information rather than global information.
We find that \sysname\ allows for more global self-attention blocks, resulting in a decreased diagonality value when compared to both Transformer and Conformer, yielding 17\% and 3\% relative reductions in average diagonality values across all layers, respectively.
\looseness=-1

\vspace{0.12cm}
\noindent \textbf{Performance on Librispeech-960h.} Table~\ref{table:res_libri_big} compares WERs of our proposed system with other architectures on the full Librispeech~\cite{librispeech} dataset. For baselines, we reuse the numbers reported by Peng et al.~\cite{e_branchformer}. We evaluate with and without the use of an external language model~(LM) during inference. In both settings, our model is on par with state-of-the-art architectures achieving comparable performance when trained with large amounts of data.
\looseness=-1

\vspace{0.12cm}
\noindent \textbf{SLU experiments.} To evaluate the effectiveness of our proposed approach on tasks other than ASR, in Table~\ref{table:res_slu} we evaluate our method on SLU with the SLURP~\cite{slurp} dataset. \sysname\ outperforms both Conformer and E-Branchformer achieving the best accuracy and F1-score for both intent classification and entity recognition tasks, while having the least number of parameters. 
\looseness=-1

\vspace{0.12cm}
\noindent \textbf{Summary of results.} In small-scale training data scenarios for ASR (Librispeech-100h, Tedlium2 and AISHELL shown in Table~\ref{table:res_main}) and SLU (SLURP), \sysname~performs consistently better than Conformer. Moreover, our proposed system exhibits comparable or improved performance when compared to other state-of-the-art Conformer variants such as CgMLP and E-Branchformer. With large scale datasets like Librispeech-960h, we find \sysname\ to be on par with all these architectures.

\begin{figure}[t!]

\centering
\includegraphics[keepaspectratio, width=0.465\textwidth]{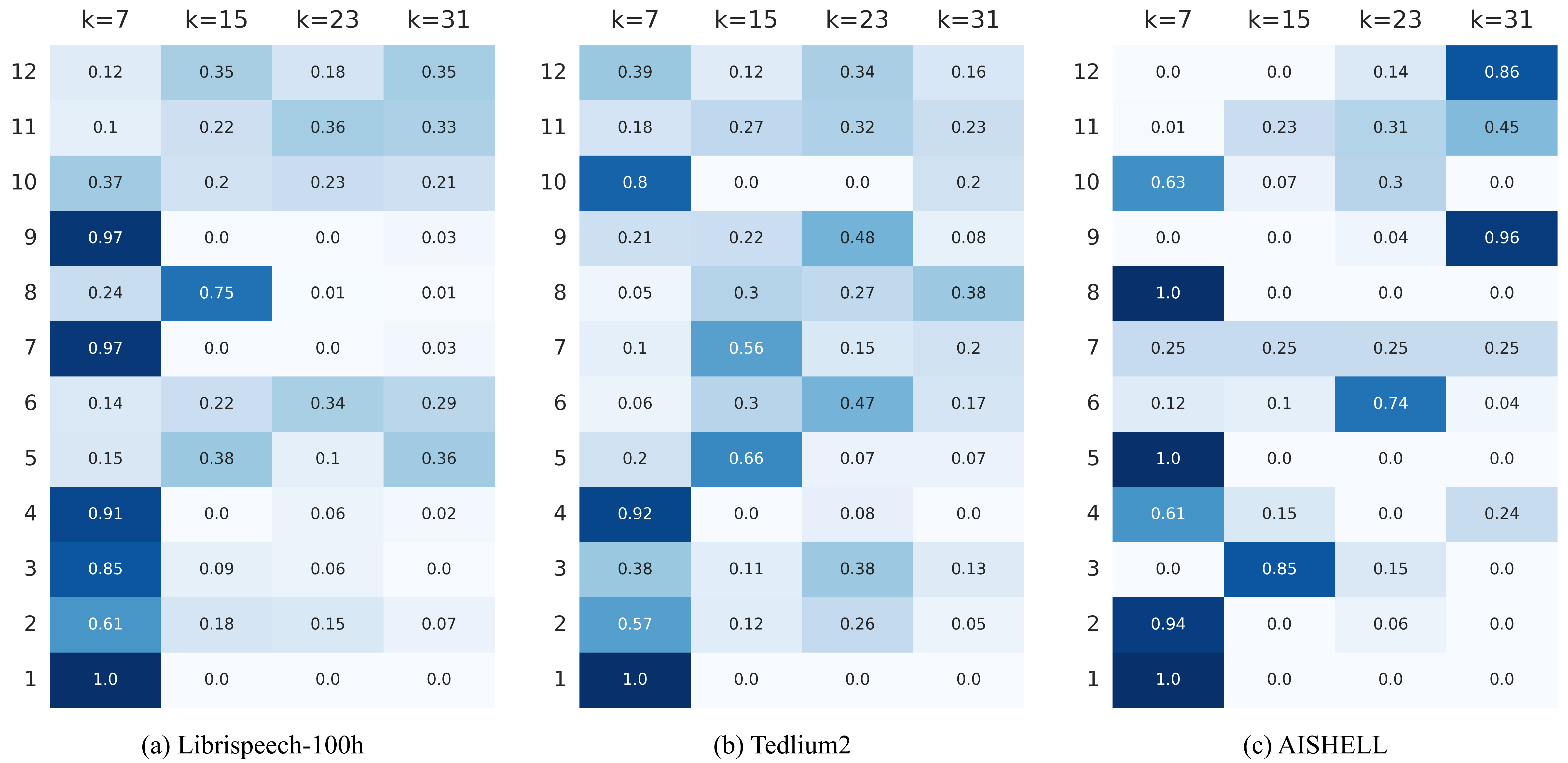}
\caption{Heatmap showing the importance given to each convolution kernel across all encoder layers. A dark blue cell  indicates a high level of importance given to a specific kernel.} 
\label{fig:analyis:gate} 
  \vspace{-0.1cm}
\end{figure}

\begin{table}[t!]
\caption{Comparison of the performance~(WER \%) of our system against other architectures on the full Librispeech dataset.} 
  \vspace{-0.05cm}

\label{table:res_libri_big}
\centering
\resizebox{\linewidth}{!}{
\begin{tabular}{ | l | c | cc | cc | }
    \hline
    \multicolumn{1}{|c|}{\multirow{2}{*}{\textbf{\small{Kernels}}}} & \multicolumn{1}{|c|}{\multirow{2}{*}{\textbf{\small{Params}}}} & \multicolumn{2}{c|}{\textbf{\footnotesize{Without LM}}} & \multicolumn{2}{c|}{\textbf{\footnotesize{With LM}}} \\
    
    \cline{3-6}
    
    &  & \textbf{\footnotesize{Test Clean}} & \textbf{\footnotesize{Test Other}} & \textbf{\footnotesize{Test Clean}} & \textbf{\footnotesize{Test Other}} \\

    \hline

   \small{Conformer}~\cite{conformer} & \footnotesize{147.8M} & \small{2.16} & \small{4.74} & \textbf{\small{1.84}} & \small{3.95} \\
   \small{Branchformer}~\cite{branchformer} & \footnotesize{146.7M} & \small{2.25} & \small{4.83} & \small{1.93} & \small{4.00} \\
   \small{E-Branchformer}~\cite{e_branchformer}  & \footnotesize{148.9M} & \textbf{\small{2.14}} & \textbf{\small{4.55}} & \small{1.85} & \textbf{\small{3.71}} \\

    \arrayrulecolor{black!50}
    \hline
    \arrayrulecolor{black}
   
   \footnotesize{$\blockname_{\texttt{fusion}}$} & \footnotesize{147.4M} & \small{2.15} & \small{4.69} & \small{1.89} & \small{3.92} \\
    
    \hline
\end{tabular}
}
\vspace{-0.35cm}
\end{table}

\begin{table}[t!]
\caption{Comparison~(accuracy \% and F1) of our system against other architectures on the SLU task.}
\label{table:res_slu}
  \vspace{-0.05cm}

\centering
\resizebox{\linewidth}{!}{
\begin{tabular}{ | l | c | cc | ccc | }
    \hline
    \multicolumn{1}{|c|}{\multirow{2}{*}{\textbf{\small{Method}}}} & \multicolumn{1}{|c|}{\multirow{2}{*}{\textbf{\small{Params}}}} & \multicolumn{2}{c|}{\textbf{\footnotesize{Intent Classification}}} & \multicolumn{3}{c|}{\textbf{\footnotesize{Entity Recognition}}} \\
    
    \cline{3-7}
    
    &  & \textbf{\footnotesize{Valid Acc.}} & \textbf{\footnotesize{Test Acc.}} & \textbf{\footnotesize{SLU-F1}} &
    \textbf{\footnotesize{Precision}}  & \textbf{\footnotesize{Recall}} \\

    \hline

   \small{Conformer}~\cite{conformer} & \footnotesize{109M} & \small{87.4} & \small{86.7} & \small{77.2} & \small{80.0} & \small{74.5} \\

   \small{E-Branchformer}~\cite{e_branchformer} & \footnotesize{110M} &  \small{88.3} & \small{\textbf{87.4}} & \small{78.5} & \small{80.7} & \small{76.5} \\

   \arrayrulecolor{black!50}
    \hline
    \arrayrulecolor{black}

   \footnotesize{$\blockname_{\texttt{fusion}}$} & \footnotesize{108M} &  \small{\textbf{88.8}} & \small{\textbf{87.4}} & \small{\textbf{78.9}} & \small{\textbf{80.8}} & \small{\textbf{77.1}} \\
    
    \hline
\end{tabular}
}
\vspace{-0.35cm}
\end{table}

\subsection{Analysis of Convolution Kernels}\label{sec:analysis:gate}

To better understand how the kernels are being utilized, in Figure~\ref{fig:analyis:gate}, we visualize the weights learned by the gate in $\blockname_{\texttt{weighted}}$~(shown in Figure~\ref{fig:arch:fusion}(b)) on the test-other split of Librispeech-100h, Tedlium2, and AISHELL datasets. 
We observe that not every layer benefits from having multiple convolution kernels. While the initial few encoder layers rely mostly on the smaller kernels, the intermediate and final encoder layers reap the most benefits of having multiple convolution kernels.
\looseness=-1
\begin{table}[t!]
\caption{Comparison of the performance~(WER \%) of our system with varying number and size of convolution kernels. }
\label{table:res_kernels}
\centering
\resizebox{\linewidth}{!}{
\begin{tabular}{ | l | c | c | c | }
    \hline
    \multicolumn{1}{|c|}{\multirow{2}{*}{\textbf{\small{Kernels}}}} & \multicolumn{1}{|c|}{\multirow{2}{*}{\textbf{\small{Params}}}} & \multicolumn{2}{c|}{\textbf{\footnotesize{Librispeech-100h}}} \\
    
    \cline{3-4}
    
    &  & \textbf{\footnotesize{Test Clean}} & \textbf{\footnotesize{Test Other}} \\

    \hline

   \small{$K=\{3,7\}$} & \footnotesize{36.8M} & \small{\textbf{6.15}} & \small{17.03} \\
   \small{$K=\{23,31\}$} & \footnotesize{37.0M} & \small{6.21} & \small{17.06} \\
   \small{$K=\{7,31\}$} & \footnotesize{36.9M} & \small{6.36} & \small{17.19} \\
   \small{$K=\{7,15,23,31\}$} & \footnotesize{37.2M} & \small{6.18} & \small{\textbf{17.00}} \\
   \small{$K=\{7,15,31,63\}$} & \footnotesize{37.4M} & \small{6.38} & \small{17.18} \\
   \small{$K=\{37,43,49,55\}$} & \footnotesize{37.8M} & \small{6.22} & \small{17.13} \\
    
    \hline
\end{tabular}
}
\vspace{-0.35cm}
\end{table}


In Table~\ref{table:res_kernels}, we evaluate the performance of our system by varying the number and size of the convolution kernels. We observe that when convolutions with large kernel sizes~(i.e. greater than $32$) are used, performance is negatively impacted. Performance also diminishes when there is a large gap between the chosen kernel sizes. Finally, using four kernels instead of two further improves performance. Thus in all our experiments, we use kernel sizes $K=\{7,15,23,31\}$.


\section{Conclusion}
In this work, we propose \sysname\ that utilizes multiple convolution kernels instead of a single fixed convolution as in Conformers. We demonstrate the effectiveness of \sysname\ by comparing with Conformer and its variants on multiple datasets~(Librispeech-960, Tedlium2, AISHELL and Librispeech-100), diverse modelling paradigms~(AED, CTC, RNN-T) and different speech tasks~(ASR, SLU). We also conduct ablations and analysis for a more comprehensive understanding of our architecture.

\section{Acknowledgements}
Our computing resources are supported by PSC Bridges2 and NCSA Delta via ACCESS allocation CIS210014, under National Science Foundation grants \#2138259, \#2138286, \#2138307, \#2137603, and \#2138296. Additionally, the third author would like to gratefully acknowledge support from the National Language Translation Mission~(NLTM): Bhashini project funded by the Ministry of Electronics and Information Technology~(MeitY), Government of India.

\section{References}
\printbibliography

@inproceedings{ctc,
author = {Graves, Alex and Fern\'{a}ndez, Santiago and Gomez, Faustino and Schmidhuber, J\"{u}rgen},
title = {Connectionist temporal classification: labelling unsegmented sequence data with recurrent neural networks},
year = {2006},
isbn = {1595933832},
publisher = {Association for Computing Machinery},
address = {New York, NY, USA},
url = {https://doi.org/10.1145/1143844.1143891},
doi = {10.1145/1143844.1143891},
booktitle = {Proceedings of the 23rd International Conference on Machine Learning},
pages = {369–376},
numpages = {8},
location = {Pittsburgh, Pennsylvania, USA}
}

@inproceedings{rnnt,
      title={Sequence Transduction with Recurrent Neural Networks}, 
      author={Alex Graves},
      year={2012},
      booktitle={proc. ICML}
}

@INPROCEEDINGS{jointctcatt,
  author={Kim, Suyoun and Hori, Takaaki and Watanabe, Shinji},
  booktitle={2017 IEEE International Conference on Acoustics, Speech and Signal Processing (ICASSP)}, 
  title={Joint CTC-attention based end-to-end speech recognition using multi-task learning}, 
  year={2017},
  volume={},
  number={},
  pages={4835-4839},
  doi={10.1109/ICASSP.2017.7953075}}

@inproceedings{attention,
 author = {Vaswani, Ashish and Shazeer, Noam and Parmar, Niki and Uszkoreit, Jakob and Jones, Llion and Gomez, Aidan N and Kaiser, \L ukasz and Polosukhin, Illia},
 booktitle = {Advances in Neural Information Processing Systems},
 pages = {},
 publisher = {Curran Associates, Inc.},
 title = {Attention is All you Need},
 volume = {30},
 year = {2017}
}

@inproceedings{gated_mlp,
 author = {Liu, Hanxiao and Dai, Zihang and So, David and Le, Quoc V},
 booktitle = {Advances in Neural Information Processing Systems},
 pages = {9204--9215},
 publisher = {Curran Associates, Inc.},
 title = {Pay Attention to MLPs},
 volume = {34},
 year = {2021}
}

@inproceedings{convolution,
 author = {Krizhevsky, Alex and Sutskever, Ilya and Hinton, Geoffrey E},
 booktitle = {Advances in Neural Information Processing Systems},
 pages = {},
 publisher = {Curran Associates, Inc.},
 title = {ImageNet Classification with Deep Convolutional Neural Networks},
 volume = {25},
 year = {2012}
}

@inproceedings{conformer,
  author={Anmol Gulati and James Qin and Chung-Cheng Chiu and Niki Parmar and Yu Zhang and Jiahui Yu and Wei Han and Shibo Wang and Zhengdong Zhang and Yonghui Wu and Ruoming Pang},
  title={{Conformer: Convolution-augmented Transformer for Speech Recognition}},
  year=2020,
  booktitle={Proc. Interspeech 2020},
  pages={5036--5040},
  doi={10.21437/Interspeech.2020-3015}
}

@article{cgmlp,
  title={MLP-ASR: Sequence-length agnostic all-MLP architectures for speech recognition},
  author={Sakuma, Jin and Komatsu, Tatsuya and Scheibler, Robin},
  journal={arXiv preprint arXiv:2202.08456},
  year={2022}
}

@InProceedings{branchformer,
  title = 	 {Branchformer: Parallel {MLP}-Attention Architectures to Capture Local and Global Context for Speech Recognition and Understanding},
  author =       {Peng, Yifan and Dalmia, Siddharth and Lane, Ian and Watanabe, Shinji},
  booktitle = 	 {Proceedings of the 39th International Conference on Machine Learning},
  pages = 	 {17627--17643},
  year = 	 {2022},
  volume = 	 {162},
  month = 	 {17--23 Jul},
  publisher =    {PMLR},
}

@INPROCEEDINGS{e_branchformer,
  author={Kim, Kwangyoun and Wu, Felix and Peng, Yifan and Pan, Jing and Sridhar, Prashant and Han, Kyu J. and Watanabe, Shinji},
  booktitle={2022 IEEE Spoken Language Technology Workshop (SLT)}, 
  title={E-Branchformer: Branchformer with Enhanced Merging for Speech Recognition}, 
  year={2023},
  volume={},
  number={},
  pages={84-91},}

@inproceedings{squeezeformer,
 author = {Kim, Sehoon and Gholami, Amir and Shaw, Albert and Lee, Nicholas and Mangalam, Karttikeya and Malik, Jitendra and Mahoney, Michael W and Keutzer, Kurt},
 booktitle = {Advances in Neural Information Processing Systems},
 pages = {9361--9373},
 publisher = {Curran Associates, Inc.},
 title = {Squeezeformer: An Efficient Transformer for Automatic Speech Recognition},
 volume = {35},
 year = {2022}
}

@article{gelu,
  title={Gaussian Error Linear Units (GELUs)},
  author={Hendrycks, Dan and Gimpel, Kevin},
  journal={arXiv preprint arXiv:1606.08415},
  year={2016}
}

@INPROCEEDINGS{speech-transformer,
  author={Dong, Linhao and Xu, Shuang and Xu, Bo},
  booktitle={2018 IEEE International Conference on Acoustics, Speech and Signal Processing (ICASSP)}, 
  title={Speech-Transformer: A No-Recurrence Sequence-to-Sequence Model for Speech Recognition}, 
  year={2018},
  volume={},
  number={},
  pages={5884-5888},
  doi={10.1109/ICASSP.2018.8462506}}

@INPROCEEDINGS{librispeech,
  author={Panayotov, Vassil and Chen, Guoguo and Povey, Daniel and Khudanpur, Sanjeev},
  booktitle={2015 IEEE International Conference on Acoustics, Speech and Signal Processing (ICASSP)}, 
  title={Librispeech: An ASR corpus based on public domain audio books}, 
  year={2015},
  volume={},
  number={},
  pages={5206-5210},
  doi={10.1109/ICASSP.2015.7178964}}

@inproceedings{aishell,
  author = {Hui, Bu and Jiayu, Du and Xingyu, Na and Bengu, Wu and Hao Zheng},
  title = {AIShell-1: An Open-Source Mandarin Speech Corpus and A Speech Recognition Baseline},
  booktitle = {Oriental COCOSDA},
  pages = {1-5},
  publisher = {{IEEE}},
  year = {2017},
  url = {https://doi.org/10.1109/ICSDA.2017.8384449},
  doi = {10.1109/ICSDA.2017.8384449},
  bibsource = {dblp computer science bibliography, https://dblp.org}
}

@inproceedings{tedlium2,
  title={Enhancing the TED-LIUM corpus with selected data for language modeling and more TED talks.},
  author={Rousseau, Anthony and Del{\'e}glise, Paul and Esteve, Yannick and others},
  booktitle={Proc. LREC},
  pages={3935--3939},
  year={2014}
}

@inproceedings{slurp,
    title = "{SLURP}: A Spoken Language Understanding Resource Package",
    author = "Bastianelli, Emanuele  and
      Vanzo, Andrea  and
      Swietojanski, Pawel  and
      Rieser, Verena",
    booktitle = "Proceedings of the 2020 Conference on Empirical Methods in Natural Language Processing (EMNLP)",
    month = nov,
    year = "2020",
    address = "Online",
    publisher = "Association for Computational Linguistics",
    doi = "10.18653/v1/2020.emnlp-main.588",
    pages = "7252--7262",
}

@inproceedings{espnet,
  author={Shinji Watanabe and Takaaki Hori and Shigeki Karita and Tomoki Hayashi and Jiro Nishitoba and Yuya Unno and Nelson {Enrique Yalta Soplin} and Jahn Heymann and Matthew Wiesner and Nanxin Chen and Adithya Renduchintala and Tsubasa Ochiai},
  title={{ESPnet}: End-to-End Speech Processing Toolkit},
  year={2018},
  booktitle={Proceedings of Interspeech},
  pages={2207--2211},
  doi={10.21437/Interspeech.2018-1456},
  url={http://dx.doi.org/10.21437/Interspeech.2018-1456}
}

@inproceedings{specaug,
  author={Daniel S. Park and William Chan and Yu Zhang and Chung-Cheng Chiu and Barret Zoph and Ekin D. Cubuk and Quoc V. Le},
  title={{SpecAugment: A Simple Data Augmentation Method for Automatic Speech Recognition}},
  year=2019,
  booktitle={Proc. Interspeech 2019},
  pages={2613--2617},
  doi={10.21437/Interspeech.2019-2680}
}

@INPROCEEDINGS{mapsswe,
  author={Gillick, L. and Cox, S.J.},
  booktitle={International Conference on Acoustics, Speech, and Signal Processing,}, 
  title={Some statistical issues in the comparison of speech recognition algorithms}, 
  year={1989},
  volume={},
  number={},
  pages={532-535 vol.1},
  doi={10.1109/ICASSP.1989.266481}}

@inproceedings{
zipformer,
title={Zipformer: A faster and better encoder for automatic speech recognition},
author={Zengwei Yao and Liyong Guo and Xiaoyu Yang and Wei Kang and Fangjun Kuang and Yifan Yang and Zengrui Jin and Long Lin and Daniel Povey},
booktitle={The Twelfth International Conference on Learning Representations},
year={2024},
url={https://openreview.net/forum?id=9WD9KwssyT}
}

@ARTICLE{leformer,
  author={Wei, Guangyong and Duan, Zhikui and Li, Shiren and Yu, Xinmei and Yang, Guangguang},
  journal={IEEE Signal Processing Letters}, 
  title={LFEformer: Local Feature Enhancement Using Sliding Window With Deformability for Automatic Speech Recognition}, 
  year={2023},
  volume={30},
  number={},
  pages={180-184},
  doi={10.1109/LSP.2023.3241558}}

@INPROCEEDINGS{dynamic_convolution,
  author={Chen, Yinpeng and Dai, Xiyang and Liu, Mengchen and Chen, Dongdong and Yuan, Lu and Liu, Zicheng},
  booktitle={Proc. CVPR}, 
  title={Dynamic Convolution: Attention Over Convolution Kernels}, 
  year={2020},
  volume={},
  number={},
  pages={11027-11036},
  doi={10.1109/CVPR42600.2020.01104}}

@InProceedings{dynamic_convolution2,
    author    = {Dai, Xiyang and Chen, Yinpeng and Yang, Jianwei and Zhang, Pengchuan and Yuan, Lu and Zhang, Lei},
    title     = {Dynamic DETR: End-to-End Object Detection With Dynamic Attention},
    booktitle = {Proc. ICCV},
    month     = {October},
    year      = {2021},
    pages     = {2988-2997}
}

@article{dynamic_convolution3,
  title={Attention in attention network for image super-resolution},
  author={Chen, Haoyu and Gu, Jinjin and Zhang, Zhi},
  journal={arXiv preprint arXiv:2104.09497},
  year={2021}
}

@article{dynamic_convolution4,
  title={Dynet: Dynamic convolution for accelerating convolutional neural networks},
  author={Zhang, Yikang and Zhang, Jian and Wang, Qiang and Zhong, Zhao},
  journal={arXiv preprint arXiv:2004.10694},
  year={2020}
}

@inproceedings{
dynamic_convolution5,
title={Revisiting Dynamic Convolution via Matrix Decomposition},
author={Yunsheng Li and Yinpeng Chen and Xiyang Dai and mengchen liu and Dongdong Chen and Ye Yu and Lu Yuan and Zicheng Liu and Mei Chen and Nuno Vasconcelos},
booktitle={Proc. ICLR},
year={2021},
url={https://openreview.net/forum?id=YwpZmcAehZ}
}

@INPROCEEDINGS{attnetion_usefullness,
  author={Zhang, Shucong and Loweimi, Erfan and Bell, Peter and Renals, Steve},
  booktitle={2021 IEEE Spoken Language Technology Workshop (SLT)}, 
  title={On The Usefulness of Self-Attention for Automatic Speech Recognition with Transformers}, 
  year={2021},
  volume={},
  number={},
  pages={89-96},
  doi={10.1109/SLT48900.2021.9383521}}

@inproceedings{comparative_study_branchformer,
  author={Yifan Peng and Kwangyoun Kim and Felix Wu and others},
  title={{A Comparative Study on E-Branchformer vs Conformer in Speech Recognition, Translation, and Understanding Tasks}},
  year=2023,
  booktitle={Proc. Interspeech 2023},
  pages={2208--2212},
  doi={10.21437/Interspeech.2023-1194}
}

@inproceedings{peng2024owsmctc,
  author={Yifan Peng and Kwangyoun Kim and Felix Wu and others},
  title={{OWSM-CTC: An Open Encoder-Only Speech Foundation Model for Speech Recognition, Translation, and Language Identification}},
  year=2024,
  booktitle={Proc. ACL},
}

@article{e2e_asr_survey1,
url = {http://dx.doi.org/10.1561/116.00000050},
year = {2022},
volume = {11},
journal = {APSIPA Transactions on Signal and Information Processing},
title = {Recent Advances in End-to-End Automatic Speech Recognition},
doi = {10.1561/116.00000050},
issn = {},
number = {1},
pages = {},
author = {Jinyu Li}
}

@article{e2e_asr_survey2,
title = "End-to-End Speech Recognition: A Survey",
author = "Rohit Prabhavalkar and Takaaki Hori and Sainath, {Tara N.} and Ralf Schluter and Shinji Watanabe",
year = "2024",
doi = "10.1109/TASLP.2023.3328283",
language = "English",
volume = "32",
pages = "325--351",
journal = "IEEE/ACM Transactions on Audio Speech and Language Processing",
issn = "2329-9290",
publisher = "IEEE Advancing Technology for Humanity",
}

@INPROCEEDINGS{conformer_impl,
  author={Guo, Pengcheng and Boyer, Florian and Chang, Xuankai and Hayashi, Tomoki and Higuchi, Yosuke and Inaguma, Hirofumi and Kamo, Naoyuki and Li, Chenda and Garcia-Romero, Daniel and Shi, Jiatong and Shi, Jing and Watanabe, Shinji and Wei, Kun and Zhang, Wangyou and Zhang, Yuekai},
  booktitle={ICASSP 2021 - 2021 IEEE International Conference on Acoustics, Speech and Signal Processing (ICASSP)}, 
  title={Recent Developments on Espnet Toolkit Boosted By Conformer}, 
  year={2021},
  volume={},
  number={},
  pages={5874-5878},
  doi={10.1109/ICASSP39728.2021.9414858}}

@article{convolution2,
  title={A guide to convolution arithmetic for deep learning},
  author={Dumoulin, Vincent and Visin, Francesco},
  journal={arXiv preprint arXiv:1603.07285},
  year={2016}
}

@INPROCEEDINGS{convolution3,
  author={Albawi, Saad and Mohammed, Tareq Abed and Al-Zawi, Saad},
  booktitle={Proc. ICET}, 
  title={Understanding of a convolutional neural network}, 
  year={2017},
  volume={},
  number={},
  pages={1-6},
  doi={10.1109/ICEngTechnol.2017.8308186}}

@INPROCEEDINGS{speech_transformer2,
  author={Dong, Linhao and Xu, Shuang and Xu, Bo},
  booktitle={2018 IEEE International Conference on Acoustics, Speech and Signal Processing (ICASSP)}, 
  title={Speech-Transformer: A No-Recurrence Sequence-to-Sequence Model for Speech Recognition}, 
  year={2018},
  volume={},
  number={},
  pages={5884-5888},
  doi={10.1109/ICASSP.2018.8462506}}

@INPROCEEDINGS{speech_transformer3,
  author={Karita, Shigeki and Chen, Nanxin and Hayashi, Tomoki and Hori, Takaaki and Inaguma, Hirofumi and Jiang, Ziyan and Someki, Masao and Soplin, Nelson Enrique Yalta and Yamamoto, Ryuichi and Wang, Xiaofei and Watanabe, Shinji and Yoshimura, Takenori and Zhang, Wangyou},
  booktitle={2019 IEEE Automatic Speech Recognition and Understanding Workshop (ASRU)}, 
  title={A Comparative Study on Transformer vs RNN in Speech Applications}, 
  year={2019},
  volume={},
  number={},
  pages={449-456},
  doi={10.1109/ASRU46091.2019.9003750}}

@inproceedings{positionwise_ff,
    title = "Transformer Feed-Forward Layers Are Key-Value Memories",
    author = "Geva, Mor  and
      Schuster, Roei  and
      Berant, Jonathan  and
      Levy, Omer",
    booktitle = "Proceedings of the 2021 Conference on Empirical Methods in Natural Language Processing",
    month = nov,
    year = "2021",
    address = "Online and Punta Cana, Dominican Republic",
    publisher = "Association for Computational Linguistics",
    url = "https://aclanthology.org/2021.emnlp-main.446",
    doi = "10.18653/v1/2021.emnlp-main.446",
    pages = "5484--5495",
}

@article{layernorm,
  title={Layer normalization},
  author={Ba, Jimmy Lei and Kiros, Jamie Ryan and Hinton, Geoffrey E},
  journal={arXiv preprint arXiv:1607.06450},
  year={2016}
}

@INPROCEEDINGS{ser,
  author={Peng, Zixuan and Lu, Yu and Pan, Shengfeng and Liu, Yunfeng},
  booktitle={ICASSP 2021 - 2021 IEEE International Conference on Acoustics, Speech and Signal Processing (ICASSP)}, 
  title={Efficient Speech Emotion Recognition Using Multi-Scale CNN and Attention}, 
  year={2021},
  volume={},
  number={},
  pages={3020-3024},
  doi={10.1109/ICASSP39728.2021.9414286}}

@INPROCEEDINGS{multi_octave,
  author={Rownicka, Joanna and Bell, Peter and Renals, Steve},
  booktitle={ICASSP 2020 - 2020 IEEE International Conference on Acoustics, Speech and Signal Processing (ICASSP)}, 
  title={Multi-Scale Octave Convolutions for Robust Speech Recognition}, 
  year={2020},
  volume={},
  number={},
  pages={7019-7023},
  doi={10.1109/ICASSP40776.2020.9053703}}

@INPROCEEDINGS{multi_stream,
  author={Han, Kyu J. and Pan, Jing and Tadala, Venkata Krishna Naveen and Ma, Tao and Povey, Dan},
  booktitle={ICASSP 2021 - 2021 IEEE International Conference on Acoustics, Speech and Signal Processing (ICASSP)}, 
  title={Multistream CNN for Robust Acoustic Modeling}, 
  year={2021},
  volume={},
  number={},
  pages={6873-6877},
  doi={10.1109/ICASSP39728.2021.9414639}}
\end{document}